\documentclass[11pt]{article}

\usepackage{authblk}

\usepackage[preprint]{acl}

\usepackage{times}
\usepackage{latexsym}

\usepackage[T1]{fontenc}

\usepackage[utf8]{inputenc}

\usepackage{microtype}

\usepackage{inconsolata}

\usepackage{graphicx}

\usepackage{enumerate}
\usepackage{amsmath} 
\usepackage[ruled,vlined]{algorithm2e}
\usepackage{booktabs}
\usepackage{multirow}
\usepackage{enumitem}
\usepackage{float}

\newcommand{\ours}{\text{CodeRAG}\xspace}

\newcommand\equalauthorfootnote[1]{%
    \begingroup
    \renewcommand\thefootnote{}\footnote{\textsuperscript{\dag}#1}%
    \addtocounter{footnote}{-1}%
    \endgroup
}

\newcommand\corrauthorfootnote[1]{%
    \begingroup
    \renewcommand\thefootnote{}\footnote{\textsuperscript{\S}#1}%
    \addtocounter{footnote}{-1}%
    \endgroup
}

\graphicspath{{figs/}}
 
\title{CodeRAG: Finding Relevant and Necessary Knowledge for Retrieval-Augmented Repository-Level Code Completion}

\author{
\textbf{Sheng Zhang\textsuperscript{1,\dag},\,\,Yifan Ding\textsuperscript{1,\dag},\,\,Shuquan Lian\textsuperscript{1},\,\,Shun Song\textsuperscript{2},\,\,Hui Li\textsuperscript{1,\S}} \\
\textsuperscript{1}Key Laboratory of Multimedia Trusted Perception and Efficient Computing\\ Ministry of Education of China, Xiamen University \\
\textsuperscript{2}Ant Group\\
\texttt{\{sheng,\,dingyf,\,shuquanlian\}@stu.xmu.edu.cn},\,\,\texttt{songshun.ss@antgroup.com}\\\texttt{hui@xmu.edu.cn}
}

\begin{document}
\maketitle
\begin{abstract}
Repository-level code completion automatically predicts the unfinished code based on the broader information from the repository. Recent strides in Code Large Language Models (code LLMs) have spurred the development of repository-level code completion methods, yielding promising results. Nevertheless, they suffer from issues such as inappropriate query construction, single-path code retrieval, and misalignment between code retriever and code LLM. To address these problems, we introduce \ours, a framework tailored to identify relevant and necessary knowledge for retrieval-augmented repository-level code completion. Its core components include log probability guided query construction, multi-path code retrieval, and preference-aligned \textsc{BestFit} reranking. Extensive experiments on benchmarks ReccEval and CCEval demonstrate that \ours significantly and consistently outperforms state-of-the-art methods. The implementation of \ours is available at \url{https://github.com/KDEGroup/CodeRAG}.
\end{abstract}

\section{Introduction}
\label{sec:intro}

Recent years\equalauthorfootnote{The first two authors contribute equally.}
\corrauthorfootnote{Hui Li is the corresponding authors.}have witnessed the remarkable success of Large Language Models (LLMs) in various areas~\cite{abs-2303-18223}. 
As a branch of LLMs, Code Large Language Models (code LLMs), are trained on massive code data, enabling them to comprehend and generate code snippets, thus assisting programmers in coding tasks and boosting development efficiency~\cite{NijkampPHTWZSX23,CodeLlama,LiAZMKMMALCLZZW23}.

A typical application of code LLMs is code completion, which automatically predicts the unfinished code~\cite{SvyatkovskiyZFS19}.
Early code completion methods solely leverage \emph{code context} (i.e., information from the function or source code file that the programmer is working on)~\cite{LiWLK18,WangL21a}. 
However, real-world software source code generally consists of multiple code files with complex interdependencies, which were neglected by early methods.
These code files are essential ingredients for programmers to consider when developing unfinished code and they are typically organized as a source code repository.
Thus, a practical code completion tool should be \emph{repository-level} and leverage both code context and information retrieved from the entire codebase to provide more accurate and comprehensive code suggestions~\cite{RepoCoder}.

Since repository-level code completion can better facilitate collaborative development and software maintenance, there is a surge of work in this direction~\cite{RepoCoder, GraphCoder, Dataflow} and most of them consider applying Retrieval-Augmented Generation (RAG), a prevalent solution incorporating external knowledge to help LLMs generate more accurate text~\cite{abs-2312-10997}. 
Based on the idea of RAG, these methods retrieve relevant code knowledge from the entire repository as supplementary to code context when predicting the unfinished code.

\begin{figure}[t]
    \centering
    \includegraphics[width=0.95\linewidth]{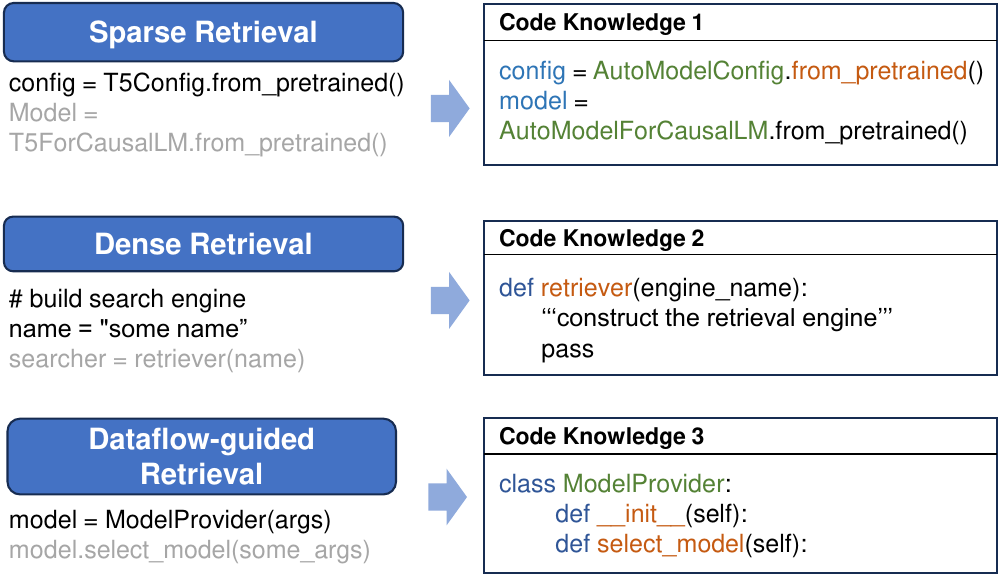}
    \caption{Examples of different code retrieval paths, where the gray text indicates the code to be generated.}
    \label{fig:ret_case}
\end{figure}

Despite the blossom of related approaches, they still suffer from the following shortcomings:
\begin{itemize}[leftmargin=10pt,topsep=1pt,itemsep=0.2pt] 

    \item \textbf{P1: Inappropriate Query Construction.} Previous approaches use either the last $k$ lines before the cursor position~\cite{abs-2405-07530} or the last $k$ lines together with the generated code from code LLM~\cite{RepoCoder} as the query for code retrieval and find relevant code knowledge to assist completion, causing information loss and introducing noises. 
    For example, programmers may define key variables and classes, or import packages at the beginning of a file, which are essential for understanding and completing the code accurately. 
    If the last $k$ lines contain irrelevant code, the retrieved code knowledge will mislead code LLM to generate inaccurate code.

    \item \textbf{P2: Single-path Code Retrieval.} Existing methods either model code as plain text and chuck code to construct the knowledge base for later sparse/dense retrieval~\cite{RepoCoder,0054AZR024}, or construct specific data structures (e.g., dataflow graph) representing code for later retrieval~\cite{GraphCoder, Dataflow}. 
    While each method has its unique advantage and may apply to some code completion cases, neither of them can handle all completion cases well. 
    For instance, Fig.~\ref{fig:ret_case} depicts three code completion examples and each fits one retrieval method. 
    Sparse retrieval is ideal when the query and code knowledge directly overlap. 
    Dense retrieval is more appropriate when the query and code knowledge are semantically related. 
    In contrast, dataflow-guided retrieval facilitates additional searches based on variable instantiation.

    \item \textbf{P3: Misalignment between Code Retriever and Code LLM.} Similar to other RAG applications~\cite{Jin00D24}, inconsistencies may exist between retrieved code knowledge and necessary knowledge for code LLM, due to the separate training process and learning objective of code retriever and code LLM. While this issue has recently been studied in various works on RAG for question answering~\cite{ZhangYWZ24,abs-2406-18676}, it is still underexplored for repository-level code completion. 
\end{itemize}

To address these issues, we propose a new framework \ours for finding \emph{relevant} and \emph{necessary} knowledge in retrieval-augmented repository-level code completion. 
Our contributions are:
\begin{itemize}[leftmargin=10pt,topsep=1pt,itemsep=0.2pt] 

\item To overcome \textbf{P1}, instead of using last $k$ lines, \ours adopts log probability guided probing to construct retrieval query for code retrieval.

\item To address \textbf{P2}, \ours employs multi-path code retrieval over the constructed code knowledge base to benefit from the unique advantage of each \emph{code-specific} retrieval path.

\item To alleviate \textbf{P3}, \ours adopts preference-aligned \textsc{BestFit} reranking to efficiently find necessary code knowledge. The retrieved code knowledge is reranked via LLM reranker according to code LLM's preference. 
To reduce the reranking overhead, we further distill the preference of LLM reranker into a smaller reranker and use it to conduct reranking.

\item \ours feeds the reranked code knowledge into code LLM for repository-level code completion. Experiments on benchmarks ReccEval and CCEval show that \ours significantly and consistently exceeds state-of-the-art methods.

\end{itemize}

\section{Our Method \ours}

\begin{figure*}[h]
    \centering
    \includegraphics[width=0.95\linewidth]{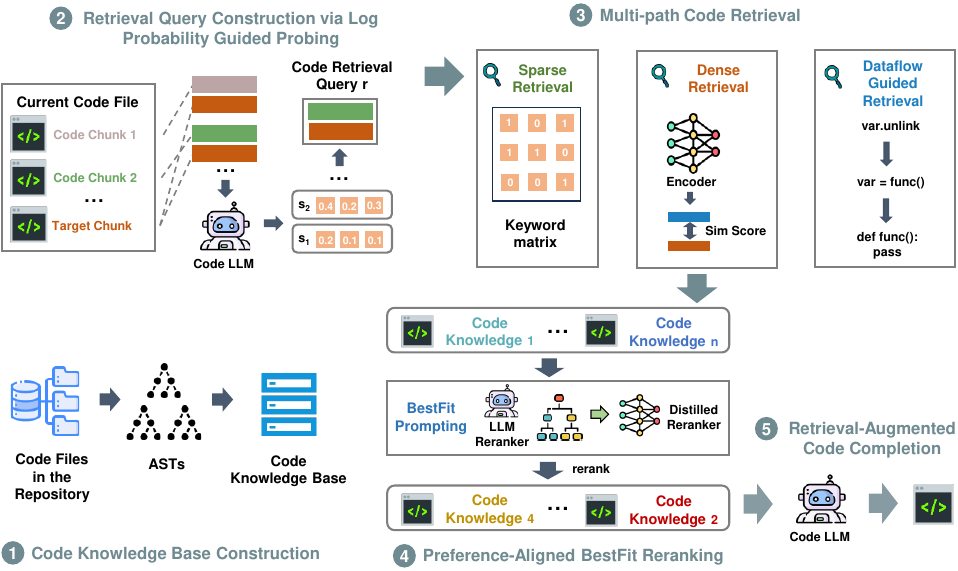}
    \caption{Overview of \ours.}
    \label{fig:overview}
\end{figure*}

As depicted in Fig.~\ref{fig:overview}, \ours involves five parts: code knowledge base construction (Sec.~\ref{sec:codekb}), retrieval query construction (Sec.~\ref{sec:query}), multi-path code retrieval (Sec.~\ref{sec:coderetrieval}), preference-aligned \textsc{BestFit} code reranking (Sec.~\ref{sec:rerank}) and retrieval-augmented repository-level code completion.

\subsection{Code Knowledge Base Construction}
\label{sec:codekb}

Constructing the repository-level code knowledge base involves parsing and processing the code in the repository, transforming raw code into structured knowledge to enable more efficient retrieval, understanding, and reuse.

\begin{figure}[h]
    \centering
    \includegraphics[width=0.68\linewidth]{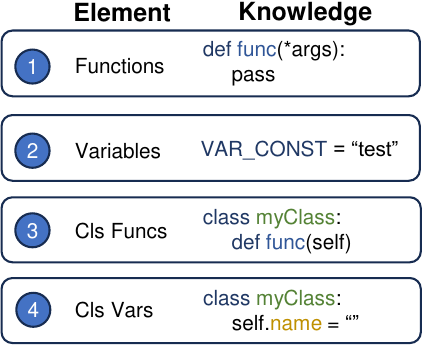}
    \caption{Examples of Code Knowledge Base Items.}
    \label{fig:code_kb_item}
\end{figure}

To construct the knowledge base for retrieval, general RAG methods typically segment the text corpus based on predefined rules, such as splitting by length or delimiters~\cite{SarthiATKGM24}.
However, applying these approaches to code data compromises the structural integrity and leads to the loss of pertinent information. 
For instance, dividing a class arbitrarily may result in omitting essential class-related details.

To alleviate the above problem, we propose a segmentation strategy tailored to the construction of code knowledge base. 
Specifically, we consider four elements in constructing the code knowledge base: functions, global variables, class variables, and class functions, as depicted in Fig.~\ref{fig:code_kb_item}. 
For a target code repository, we first extract the Abstract Syntax Tree (AST) of each code file. 
Then, we extract the four types of elements from ASTs. 
This way, the code repository can be transformed into a structured knowledge base, including function calls and variable usage, providing data support for repository-level code completion.

\subsection{Retrieval Query Construction via Log Probability Guided Probing}
\label{sec:query}

\begin{algorithm}[t]
\small
\caption{Construct Retrieval Query}
\KwIn{$C$ \text{(code file to be completed)}, $f$ (chunk length), $m$ (number of generation step), $g$ (number of selected chunks)}
\KwOut{$r$ \text{(code retrieval query)}}
\label{alg:query_by_logit}
\SetKwFunction{FMain}{QueryConstruction}
\SetKwProg{Fn}{Function}{:}{}
\Fn{\FMain{$C$, $f$, $m$, $g$}}{
    Divide $C$ into fine-grained chunks and each chunk having $f$ lines.
    
    \For{each chunk $c_i$}{

        \If{$c_i$ is the target chunk}{
            Pass.
        }
        
        Concatenate $c_i$ to the target chunk.

        Feed the concatenation into code LLM to generate $m$ new tokens.
        
        Record the highest log probability for all tokens in the vocabulary at each generation step.
        
        Sum $m$ probability scores as the confidence score $s_i$.
    }
    
    Select the top-$g$ chunks with the highest confidence scores $s$.
    
    Concatenate the $g$ chunks with the target chunk as the retrieval query $r$.
    
    \KwRet{$r$}.
}
\end{algorithm}

In standard RAG, a retrieval query conveys the user intent from the user query or consists of a specific text chunk from a document.  
Using the retrieval query, the RAG framework can retrieve relevant knowledge from the knowledge base to assist text generation. 
In existing repository-level code completion methods, the concept of retrieval query shifts to represent an incomplete code segment~\cite{RepoCoder}, which could be an unfinished function, a partially defined variable, or a method call within a class (i.e., code context). 

To overcome the limitation of using the last $k$ lines as the code retrieval query (i.e., \textbf{P1} illustrated in Sec.~\ref{sec:intro}), we propose to construct the code retrieval query based on the log probability gain.
Alg.~\ref{alg:query_by_logit} depicts the overall procedure for code retrieval query construction.
The core idea is to use log probability to find the fine-grained code chunks that are most important to constructing code retrieval query. 
In repository-level code completion, we can view log probability as the confidence of code LLMs, given the code retrieval query.  
In other words, log probability can reveal the relevance of the code chunks in the retrieval query.

As shown in Alg.~\ref{alg:query_by_logit}, we first chunk the code file that the programmer is working on into fine-grained pieces and each of them contains $f$ lines. 
Then, we concatenate each fine-grained chunk to the chunk containing unfinished code (the target chunk) as the probe and feed it to code LLM to generate $m$ tokens.  
For simplicity, we choose the token with the maximum log probability at each step and use CodeT5p-220m\footnote{\url{https://huggingface.co/Salesforce/codet5p-220m}} as code LLM for this step.
The sum of the log probabilities for all generated token is recorded as the relevance score for the fine-grained chunk corresponding to the probe.
Finally, the top-$g$ fine-grained chunks with the highest relevance scores are concatenated together with the target chunk as the retrieval query.

\subsubsection{Multi-path Code Retrieval}
\label{sec:coderetrieval}

As an essential part of RAG, code retriever finds relevant code knowledge from code knowledge base according to the code retrieval query.
Early code retrieval methods rely on traditional information retrieval methods like TF-IDF and BM25 (sparse retrieval), and recent code retrieval approaches commonly adopt embedding based methods (Dense Retrieval)~\cite{SunFGHCZGLC24}.
Most recently, \citet{Dataflow} find that dataflow can also be used to guide code retrieval (dataflow-guided retrieval).
These methods consider code retrieval from a single perspective and retrieve word-matching knowledge, semantically relevant knowledge, or knowledge having data dependency relations with the target chunk.
Each of them has its unique advantages and can well provide retrieved code knowledge for some code completion cases.
Hence, we argue that conducting a multi-path code retrieval can better offer code knowledge for later code completion.
Our designed multi-path code retrieval step involves the following three code retrieval paths:

\vspace{3pt}
\noindent\textbf{Sparse Retrieval}. 
Sparse retrieval relies on keyword matching between the retrieval query and code knowledge in the code knowledge base, which identifies exact or closely related keywords within the codebase, to obtain relevant invocation details, API calls, and code snippets that share similar structures or functionality. 
Sparse retrieval is efficient and particularly effective when searching syntactically similar code or commonly used functions, as it can quickly pinpoint segments that contain specific terms or identifiers. 
We use TF-IDF~\cite{Jones04} for sparse retrieval.

\vspace{3pt}
\noindent\textbf{Dense Retrieval.}  
Dense retrieval leverages an encoding model to encode the retrieval query and code knowledge in the code knowledge base into representations. 
The query is encoded at the chunk level, whereas the items in the knowledge base are either at the function level (functions) or the line level (variables).
Code knowledge that has high similarity with the retrieval query w.r.t. their representations is retrieved. 
We use cosine similarity as the similarity measure and adopt the pre-trained encoder in CodeT5p-220m as the encoding model.

\vspace{3pt}
\noindent\textbf{Dataflow-Guided Retrieval}. 
It finds relevant information w.r.t. the target chunk in the current code file according to data dependency relations.
Following \citet{Dataflow}, we first formulate the unfinished code file into a dataflow graph. 
Once the graph is built, we can retrieve the dependency starting from the last unfinished line in the dataflow graph as the retrieved code knowledge. 

\vspace{3pt}
For sparse and dense retrieval, we use the constructed retrieval query to retrieve $j$ results from each path.
If data dependency exists in the dataflow graph, we retrieve dependency-related code via dataflow-guided retrieval.
After receiving all retrieved results from three paths, we add them to a retrieval list containing $n$ (i.e., $2j+1$) results.

\subsection{Find Necessary Code Knowledge through Preference-Aligned \textsc{BestFit} Reranking}
\label{sec:rerank}

The retrieved code knowledge is used to augment the code completion prompt, directly affecting the quality of code completion.
Solely using the multi-path code retriever may not provide an appropriate order of relevant knowledge.
The reason is the misalignment between the code retriever and code LLM, which is caused by their separate training objectives~\cite{ZhangYWZ24}. 
Therefore, we further deploy a reranking module that reranks retrieved code knowledge according to code LLM's preference and only keep top-$u$ code knowledge ($u < n$).

\subsubsection{\textsc{BestFit} Code Reranking}
\label{sec:align}

To address the misalignment, a natural way is to train the reranker using feedback signals from code LLM.
However, in repository-level code completion, it is very difficult to acquire feedback from code LLM that can perfectly show the quality of the generated code. 
One possible solution is applying unit tests on the generated code from code LLM~\cite{abs-2502-11460}. 
While conducting unit tests is possible for function-level code completion, it is costly in the repository-level setting where the complete project must be executed in order to see the impact of inserting generated code. 
Besides, unlike function-level code completion where inputs and outputs to unit test are easy to design, crafting inputs and labeling outputs to unit tests in the repository-level setting is much harder (e.g., more execution parameters or outputs are not variables).
 
Considering the above difficulty, an alternative is to apply an LLM as a zero-shot reranker~\cite{0001YMWRCYR23}.
And the LLM is instructed to directly produce the reranking list of the retrieval code knowledge pieces according to their relevance to the query.
Although recent studies~\cite{0001YMWRCYR23,abs-2309-15088} have shown the strong ability of LLMs on zero-shot document reranking, we empirically find that this listwise prompting solution does not work well on reranking code knowledge: 
(1) LLMs with a few billion parameters that can be deployed locally more easily do not strictly adhere to listwise prompting, while calling APIs of online LLMs that have much larger model sizes and can understand and strictly follow listwise prompting incurs high overhead. (2) Listwise prompting itself is computationally intensive since the reranking list is generated token by token. LLM reranker must do one inference for each next token prediction during reranking list generation.

\begin{figure}[t]
    \centering
    \includegraphics[width=0.95\columnwidth]{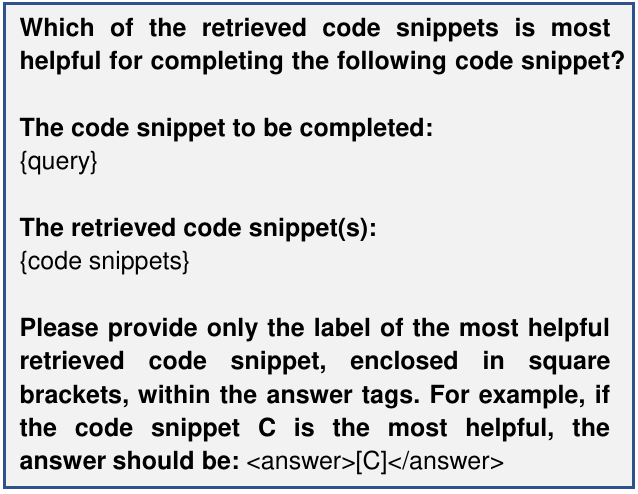}
    \caption{Prompt used for LLM-based \textsc{BestFit} code reranking.}
    \label{fig:prompt}
\end{figure}

\begin{figure}[t]
    \centering
    \includegraphics[width=1\linewidth]{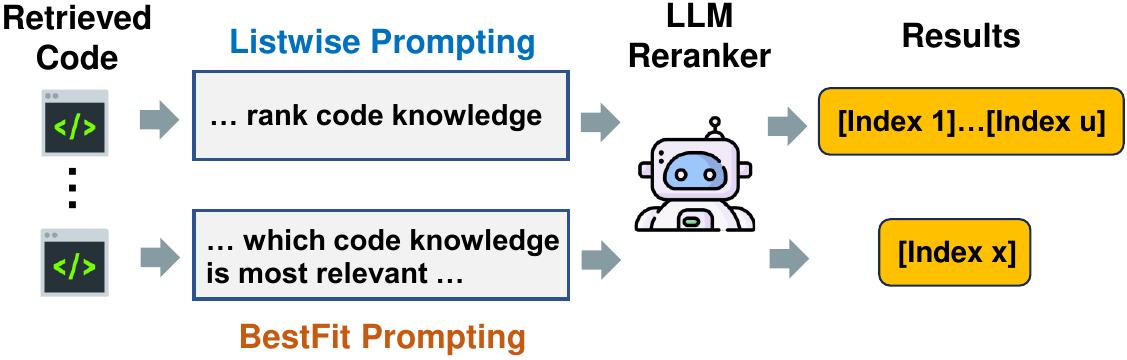}
    \caption{A comparison between listwise code reranking and \textsc{BestFit} code reranking.}
    \label{fig:bestfit1}
\end{figure}

\begin{figure}[t]
    \centering
    \includegraphics[width=0.75\linewidth]{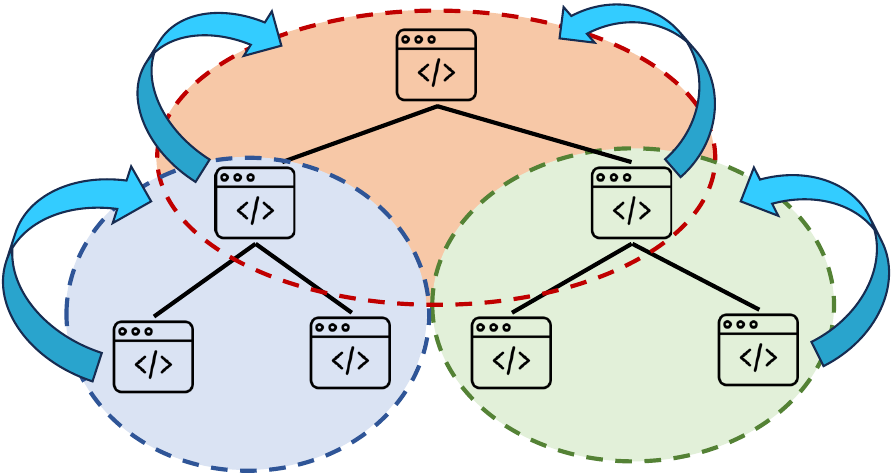}
    \caption{Heap sort operation finally moves top-$u$ code knowledge pieces to the top. Each circle denotes a window of 3 code knowledge pieces.}
    \label{fig:bestfit2}
\end{figure}

To overcome this issue, we propose \textsc{BestFit} code reranking that prompts the LLM reranker to pick the most relevant code knowledge from the retrieval list to the query. 
The prompt is listed in Fig.~\ref{fig:prompt}. 
This way, the inference cost is significantly reduced as we only need a single forward pass of the LLM reranker. Fig.~\ref{fig:bestfit1} depicts the difference between listwise and \textsc{BestFit} code reranking. Moreover, we find that an LLM with a few billion parameters can strictly follow \textsc{BestFit} prompting.  
Hence, we directly use Qwen3-8B as LLM reranker\footnote{\url{https://huggingface.co/Qwen/Qwen3-8B}}, avoiding additional instructing tuning of LLM reranker or calling online LLM APIs.

To avoid exceeding LLM's input length, we implement a sliding window strategy that divides the retrieval list into several equal-sized windows, and the adjacent two windows share one code knowledge.
Fig.~\ref{fig:bestfit2} provides an example with a window size of 3.
Each time, we feed one window to LLM reranker and ask it to pick only the most helpful code knowledge.
Inspired by prior work~\cite{QinJHZWYSLLMWB24,ZhuangZKZ24} that uses sorting algorithms to speed up LLM-based pairwise reranking, we apply heap sort to accelerate \textsc{BestFit} code reranking. 
Windows are organized as a heap tree and each time we use LLM reranker as the comparator to find the most relevant code in a window.
Heap sort can quickly find the top-$u$ most relevant code knowledge in the reranking list.
We choose heap sort instead of other sorting methods due to its simplicity and the complexity $\mathcal{O}(N logN)$.

\subsubsection{Distilled Reranker}

Even though \textsc{BestFit} code reranking only requires the LLM reranker to have a few billion parameters, directly employing the LLM reranker may still incur a high computational cost. 
Hence, we distill the preference of the LLM reranker into a much smaller reranker model. 

\begin{algorithm}[t]
\footnotesize
\caption{Construct Distillation Data}
\KwIn{$r$ \text{(code retrieval query)}, $\mathcal{L}$ \text{(initial retrieval list for $r$)}, $\mathcal{N}$ (sample numbers)}
\KwOut{$\mathcal{S}$ \text{(distillation training sample for $r$)}}
\label{alg:data}
\SetKwFunction{FMain}{DataConstruction}
\SetKwProg{Fn}{Function}{:}{}
\Fn{\FMain{$r$, $\mathcal{L}$}}{
    
    \For{$i$ in $\mathcal{N}$}{

        \For{$j = 1$ \textit{to} $3$}{
            Randomly pick $i$ code knowledge from $\mathcal{L}$ as the retrieved code snippet(s) \{code snippets\} in Fig.~\ref{fig:prompt}.

            $\mathcal{C} \leftarrow [\,\,]$
            
            \For{$z = 1$ \textit{to} $5$}{
                Use \textsc{BestFit} reranking prompt in Fig.~\ref{fig:prompt} to guide LLM reranker to select [C] from \{code snippets\}.
                Add [C] to $\mathcal{C}$.
            }

            \If{One code knowledge [C] occurs at least four times in $\mathcal{C}$}{
                Add $\left\{r, \left\{\text{code snippets}\right\}, [C]\right\}$ to $\mathcal{S}$.
            }
        }
    }

    \KwRet{$\mathcal{S}$}.
}
\end{algorithm}

To train the distilled reranker, we first use a data augmentation strategy (Alg.~\ref{alg:data}) to construct distillation training data. 
We use unfinished code in the training data to formulate code retrieval queries and conduct multi-path code retrieval to produce initial retrieval lists.
Then, we conduct data augmentation by generating multiple variations of each initial retrieval list $\mathcal{L}$, where each variation contains differing amounts of code knowledge. 
After that, we use LLM reranker with \textsc{BestFit} prompting to select the most helpful code knowledge from each generated variation, repeating this process five times to assess selection consistency. 
When LLM reranker demonstrates high confidence in its selections (i.e., when consistent choices appear in at least four out of five trials), the generated list, together with the corresponding code retrieval query and consensus selection, is treated as a distillation training sample $\mathcal{S}$.
This way, $\mathcal{S}$ reflects LLM reranker's most reliable decision patterns. 
We use all curated distillation samples to fine-tune Qwen3-0.6B\footnote{\url{https://huggingface.co/Qwen/Qwen3-0.6B}}, the backbone of the distilled reranker, using LoRA~\cite{HuSWALWWC22} and token-level cross-entropy loss.

Finally, the trained distilled reranker is used in \ours to actually rerank the retrieved code knowledge and \ours retains top-$u$ code knowledge in the reranking list ($u < n$).

\subsection{Putting All Together}

During completion, for an unfinished code file, \ours firstly constructs the corresponding code retrieval query $r$.
Then \ours uses $r$ to conduct multi-path code retrieval over the code knowledge base and retrieves top-$n$ relevant code knowledge.
After that, \ours leverages the \textsc{BestFit} reranker to rank the $n$ relevant code knowledge and retains the top-$u$ necessary code knowledge pieces.
Finally, the code context of the unfinished code file is concatenated with the $u$ pieces of code knowledge, and the result is fed into code LLM to generate the completion.

\section{Experiment}

\subsection{Evaluation Settings}

\noindent\textbf{Metrics.} 
We use prevalent metrics~\cite{GraphCoder, Dataflow, RepoCoder} to evaluate whether the generated code matches the ground-truth code (code match) and whether the identifiers in the generated code match the ground-truth identifiers (identifier match):
\begin{itemize}[leftmargin=10pt,topsep=1pt,itemsep=0.2pt] 
    \item \textbf{Code Match.} Exact Match (EM) and Edit Similarity (ES)\footnote{Note that the paper of DraCo adopts a different way to measure ES (see \url{https://github.com/nju-websoft/DraCo?tab=readme-ov-file\#evaluation}). We follow the definition of ES used in the paper of RepoCoder.} are employed to assess code alignment. 
    EM is 1 when the generated code is identical to the ground-truth answer, and 0 otherwise. 
    ES provides a more nuanced evaluation, calculated as $\text{ES} = 1 - \text{Lev}(x, y)/{max(\|x\|, \|y\|)}$, where $\text{Lev}(\cdot)$ denotes the Levenshtein distance.
    
    \item \textbf{Identifier Match}. We utilize EM and F1 scores to evaluate the alignment of identifiers in the generated code and the ground-truth answer. 
\end{itemize}

\vspace{5pt}
\noindent\textbf{Baselines.}
We use the following representative repository-level code completion baselines:
\begin{itemize}[leftmargin=10pt,topsep=1pt,itemsep=0.2pt] 
    
    \item \textbf{Zero-Shot}\footnote{\url{https://huggingface.co/Salesforce/codegen-350M-mono}\label{foot:codegen350m}}. 
    Zero-Shot directly feeds the completed code before the cursor position into code LLM without utilizing any repository information, which can reveal code LLM's basic ability to complete the code. Following existing works, we directly feed the completed code into code LLM without further using a prompt.  
     
    \item \textbf{CCFinder}\footnote{\url{https://github.com/amazon-science/cocomic}}~\cite{DingWARNBRX24}.
    CCFinder is a retrieval tool for searching repository contexts for the incomplete file. Based on import statements, it retrieves 2-hop corresponding contexts to augment completion.
     
    \item \textbf{RG-1}\footnote{\url{https://github.com/microsoft/CodeT/tree/main/RepoCoder}\label{foot:repocoder}}~\cite{RepoCoder}. 
    RG-1 is the standard RAG pipeline, which contains a Bag of Word retriever~\cite{SaltonWY75} (sparse retrieval) to retrieve the code knowledge from the code knowledge base using the unfinished code as the query. The code knowledge base consists of equal-length code chunks.

    \item \textbf{RepoCoder}\textsuperscript{\ref{foot:repocoder}}~\cite{RepoCoder}.
    RepoCoder generates code iteratively, leveraging the generation in the previous cycle to retrieve information from the code knowledge base and enhance the next generation. The code knowledge base consists of equal-length code chunks.

    \item \textbf{DraCo}\footnote{\url{https://github.com/nju-websoft/DraCo}}~\cite{Dataflow}. 
    DraCo extracts code entities and their relations through dataflow analysis, forming a repository-specific context graph. 
    During completion, it searches the graph to retrieve code features to enhance generation.  

    \item \textbf{RepoFuse}\footnote{\url{https://github.com/codefuse-ai/RepoFuse}}~\cite{abs-2402-14323}. RepoFuse~\cite{abs-2402-14323} fuses analogy context and rationale context (sparse retrieval using BM25) using a code LM to score and choose the most similar chunks to the target chunk to construct the code completion prompt. We choose UniXcoder~\cite{GuoLDW0022} as the scoring module as it shows the best result in the paper.
    
    \item \textbf{Repoformer-3B}\footnote{\url{https://huggingface.co/xiaowu0162/repoformer-3b}}~\cite{0054AZR024}.
    Repoformer-3B is a code LLM that can self-evaluate whether retrieval is necessary. It acts as both the selective RAG policy and the generation model. \emph{Note that its definition of code completion differs from other baselines and \ours since it assumes both the left part and the right part of cursor position are contexts}.
       
\end{itemize}

\vspace{5pt}
\noindent\textbf{Code LLMs.}
We use four representative code LLMs with different parameter numbers ranging from 350M to 7B as the code generators for all baselines and \ours: CodeGen-350M\textsuperscript{\ref{foot:codegen350m}}~\cite{NijkampPHTWZSX23}, SantaCoder-1.1B\footnote{\url{https://huggingface.co/bigcode/santacoder}}~\cite{abs-2301-03988}, StarCoder2-3B\footnote{\url{https://huggingface.co/bigcode/starcoder2-3b}}~\cite{abs-2402-19173}, 
and Qwen2.5-Coder-7B\footnote{\url{https://huggingface.co/Qwen/Qwen2.5-Coder-7B}}~\cite{abs-2409-12186}. 

\vspace{5pt}
\noindent\textbf{Datasets.} 
We use two benchmarks for evaluation: 
\begin{itemize}[leftmargin=10pt,topsep=1pt,itemsep=0.2pt] 
    
    \item \textbf{ReccEval}\footnote{\url{https://github.com/nju-websoft/DraCo}}~\cite{Dataflow}. ReccEval contains 6,461 test cases. It focuses on Python and contains projects first released on PyPI between 2023-01-01 and 2023-04-28. It is released under GPL-3.0 license.

    \item \textbf{CCEval}\footnote{\url{https://github.com/amazon-science/cceval}}~\cite{DingWADTJRNBRX23}. CCEval is a multilingual benchmark for repository-level code completion, where the statement to be completed has at least one use of cross-file API. CCEval contains 2,665 test cases.
    We conduct experiments on the Python subset. 
    CCEval is released under Apache-2.0 license.
    
\end{itemize}

\vspace{5pt}
\noindent\textbf{Environment and Hyper-Parameters.} 
We use a machine with two Intel(R) Xeon(R) Silver 4314 CPU @ 2.40GHz and one NVIDIA A800 GPU for experiments. 
The maximum number of generation tokens is set to 48. 
The temperature during generation is set to 0. The maximum input length of all code LLMs is set to 2,048 by default. 
We use the Text Generation Inference\footnote{\url{https://huggingface.co/docs/text-generation-inference/index}} framework to accelerate LLM inference.
By default, we use LLM reranker in \ours.
We set $f$ and $g$ to 3 and 1 in Alg.~\ref{alg:query_by_logit}, respectively. 
We set $j$ to 15 in multi-path code retrieval and $u$ to 10 in reranking. 
We use $\mathcal{N}=\{2,3,4,5,6,7\}$ in Alg.~\ref{alg:data}.

\subsection[Evaluation Results]{Evaluation Results\footnote{The results for CCEval are reported in Appendix~\ref{app:crosscodeeval}.}}

\subsubsection{Overall Performance}

\begin{table*}[t]
  \centering
  \caption{Performance on ReccEval (Use 100\% data for evaluation). Bold and underlined values indicate the best and the second-best results, respectively.}
  \scalebox{0.65}{
    \begin{tabular}{ccccccccccccccccc}
    \toprule
    \multirow{3}[2]{*}{Methods} & \multicolumn{4}{c}{CodeGen-350M} & \multicolumn{4}{c}{SantaCoder-1.1B} & \multicolumn{4}{c}{StarCoder2-3B} & \multicolumn{4}{c}{Qwen2.5-Coder-7B} \\
          & \multicolumn{2}{c}{Code Match } & \multicolumn{2}{c}{Identifier Match} & \multicolumn{2}{c}{Code Match } & \multicolumn{2}{c}{Identifier Match} & \multicolumn{2}{c}{Code Match } & \multicolumn{2}{c}{Identifier Match} & \multicolumn{2}{c}{Code Match } & \multicolumn{2}{c}{Identifier Match} \\
          & EM    & ES    & EM    & F1    & EM    & ES    & EM    & F1    & EM    & ES    & EM    & F1    & EM    & ES    & EM    & F1 \\
    \midrule
    Zero-Shot & 4.04  & 38.36  & 9.74  & 26.06  & 6.27  & 42.22  & 12.89  & 30.08  & 7.86  & 45.04  & 14.44  & 33.34  & 11.48  & 47.72  & 18.37  & 36.43  \\
    CCFinder & 16.50  & 47.71  & 23.34  & 40.12  & 19.08  & 50.99  & 26.67  & 43.31  & 28.12  & 58.93  & 36.44  & 53.42  & 28.43  & 58.95  & 36.76  & 53.09  \\
    RG-1  & 20.04  & 50.30  & 26.53  & 41.35  & 24.07  & 54.72  & 31.26  & 46.29  & 29.35  & 59.43  & 36.76  & 52.27  & 33.01  & 61.55  & 40.15  & 54.68  \\
    RepoCoder & \underline{23.96}  & \underline{53.27}  & \underline{31.01}  & 45.87  & 26.78  & 56.59  & 34.31  & 49.07  & 34.27  & 63.09  & 42.30  & 57.39  & 34.99  & 62.71  & 42.38  & 56.20  \\ 
    DraCo & 21.85  & 51.44  & 29.44  & \underline{45.92}  & \underline{30.27}  & \underline{59.38}  & \underline{38.97}  & \underline{55.50}  & \underline{36.57}  & \underline{64.31}  & \underline{45.61}  & \underline{61.42}  & \underline{39.99}  & \underline{66.26}  & \underline{48.55}  & \underline{63.41}  \\
    RepoFuse & 21.20 & 51.18 & 27.81  & 43.84 & 28.73 & 57.89 & 36.50  & 51.74 & 33.88 & 61.96 & 41.43  & 56.52  & 38.24 & 65.11 & 45.88  & 60.11 \\
    \midrule
    $\ours_{\text{llmr}}$ & \textbf{26.81}  & \textbf{55.54}  & \textbf{34.02}  & \textbf{50.13}  & \textbf{36.17}  & \textbf{63.00}  & \textbf{44.17}  & \textbf{59.64}  & \textbf{42.69}  & \textbf{68.07}  & \textbf{51.34}  & \textbf{65.73}  & \textbf{47.48}  & \textbf{70.82}  & \textbf{55.47}  & \textbf{68.68}
    \\
    \bottomrule
    \end{tabular}%
    }
    \label{tab:overall}%
\end{table*}%

\begin{table}[t]
  \centering
  \caption{Performance of RepoFormer-3B on ReccEval (Use 100\% data for evaluation). $l$: only use left context.}
  \scalebox{0.8}{
    \begin{tabular}{ccccc}
    \toprule
    \multirow{1}[1]{*}{Dataset} & \multicolumn{4}{c}{RepoFormer-3B} \\
          & \multicolumn{2}{c}{Code Match } & \multicolumn{2}{c}{Identifier Match} \\
          & EM    & ES    & EM    & F1 \\
    \midrule
    ReccEval$_{l}$ & 12.88  & 48.21  & 19.81  & 36.93  \\
    \bottomrule
    \end{tabular}%
  }
  \label{tab:repoformer_recceval_alldata}%
\end{table}%

We first report the results when using all ReccEval data for evaluation in Tab.~\ref{tab:overall} and Tab.~\ref{tab:repoformer_recceval_alldata}, as done in existing works. 
In this case, LLM reranker (i.e., $\ours_{\text{llmr}}$) is used instead of distilled reranker (i.e., $\ours_{\text{disr}}$) since distilled reranker requires training data. 
The results of splitting ReccEval for training and evaluating the distilled reranker are analyzed in Sec.~\ref{sec:comp_reranker}.
From Tab.~\ref{tab:overall} and Tab.~\ref{tab:repoformer_recceval_alldata}, we have the following observations:
\begin{itemize}[leftmargin=10pt,topsep=1pt,itemsep=0.2pt] 
    
    \item It is clear that Zero-Shot has the worst performance across all settings, showing that solely relying on code LLMs cannot provide satisfying repository-level code completion. 

    \item CCFinder and RG-1 show significant improvements over Zero-Shot but they are outperformed by more sophisticated approaches. The results indicate that RAG indeed enhances code completion but the improvements are limited since general RAG techniques are not tailored to repository-level code completion. 

    \item RepoCoder consistently outperforms CCFinder and RG-1, demonstrating that iterative retrieval~\cite{ShaoGSHDC23} exceeds naive RAG in code completion. 

    \item DraCo and RepoFuse show competitive performance and they show much better results than other baselines when larger code LLMs are used, showing that larger code LLMs may better understand the data dependence features.

    \item Comparing Tab.~\ref{tab:overall} and Tab.~\ref{tab:repoformer_recceval_alldata}, we can find that RepoFormer-3B using only left context lags behind other methods except Zero-Shot. 
    The possible reason is that RepoFormer-3B is optimized to consider both left and right contexts for code completion while other baselines and \ours are designed to consider only the left context.
    On ReccEval, where the right context is unavailable, the performance of RepoFormer-3B significantly degrades.
    We provide the results of RepoFormer-3B on CCEval, where both left and right contexts are available, in Tab.~\ref{tab:crosscodeeval_repoformer} of Appendix~\ref{app:crosscodeeval}.
    
    \item $\ours$ achieves the best performance across all settings on ReccEval, showing its effectiveness.  
    Besides, the performance gains over baselines consistently exist as the size of code LLM increases, showing the robustness of \ours. 
    
\end{itemize}

\subsubsection{Ablation Study}

\begin{table*}[t]
  \centering
  \caption{Comparisons among different variations of CodeRAG on ReccEval (Use 100\% data for evaluation). Bold and underlined values indicate the best and the second-best results, respectively.}
  \scalebox{0.63}{
    \begin{tabular}{ccccccccccccccccc}
    \toprule
    \multirow{3}[2]{*}{Methods} & \multicolumn{4}{c}{CodeGen-350M} & \multicolumn{4}{c}{SantaCoder-1.1B} & \multicolumn{4}{c}{StarCoder2-3B} & \multicolumn{4}{c}{Qwen2.5-Coder-7B} \\
          & \multicolumn{2}{c}{Code Match } & \multicolumn{2}{c}{Identifier Match} & \multicolumn{2}{c}{Code Match } & \multicolumn{2}{c}{Identifier Match} & \multicolumn{2}{c}{Code Match } & \multicolumn{2}{c}{Identifier Match} & \multicolumn{2}{c}{Code Match } & \multicolumn{2}{c}{Identifier Match} \\
          & EM    & ES    & EM    & F1    & EM    & ES    & EM    & F1    & EM    & ES    & EM    & F1    & EM    & ES    & EM    & F1 \\
    \midrule
    $\ours_{\text{s}}$ & 22.07  & 51.31  & 29.03  & 44.80  & 23.95 & 56.37  & 31.36  & 47.71  & 33.83  & 62.03  & 41.95  & 57.18  & 39.89  & 59.46  & 36.76  & 61.24  \\
    $\ours_{\text{d}}$ & 20.49  & 50.36  & 27.33  & 43.41  & 23.04  & 55.58  & 30.52  & 46.90  & 32.81  & 61.53  & 41.18  & 56.55  & 39.66  & 58.47  & 36.05  & 60.90  \\
    $\ours_{\text{d+s}}$ & 22.88  & 51.95  & 29.42  & 45.51  & 24.90  & 57.23  & 32.38  & 48.58  & 35.27  & 63.23  & 43.80  & 58.96  & 41.95  & 59.59  & 37.95  & 62.14  \\
    $\ours_{\text{df}}$ & 21.28  & 50.94  & 28.82  & 45.20  & 25.68  & 57.38  & 33.98  & 50.99  & 35.13  & 63.25  & 43.91  & 60.15  & 41.86  & 59.06  & 37.61  & 62.79  \\
    $\ours_{\text{df+s}}$ & \underline{24.21}  & \underline{53.01}  & \underline{31.93}  & \underline{47.94}  & 27.18  & 59.52  & 35.47  & 52.10  & 39.85  & 66.35  & 48.72  & 64.01  & 44.21  & 62.36  & 52.80  & \underline{66.62}  \\
    $\ours_{\text{df+s+d}}$ & 23.89 & 52.86  & 31.56  & 47.80  & \underline{27.70}  & \underline{59.96} & \underline{36.10}  & \underline{52.72}  & \underline{40.52}  & \underline{66.87}  & \underline{49.65}  & \underline{64.83}  & \underline{44.61}  & \underline{68.81}  & \underline{53.75}  & 66.57  \\
    $\ours_{\text{df+s+d+lr}}$ & \textbf{26.81}  & \textbf{55.54}  & \textbf{34.02}  & \textbf{50.13}  & \textbf{36.17}  & \textbf{63.00}  & \textbf{44.17}  & \textbf{59.64}  & \textbf{42.69}  & \textbf{68.07}  & \textbf{51.34}  & \textbf{65.73}  & \textbf{47.48}  & \textbf{70.82}  & \textbf{55.47}  & \textbf{68.68} \\
    \bottomrule
    \end{tabular}%
    }
  \label{tab:module_ablation}%
\end{table*}%

To investigate whether each part in \ours indeed contributes to its performance, we compare \ours with its variations on ReccEval using 100\% data for evaluation. 
The results are reported in Tab.~\ref{tab:module_ablation}.
Subscripts ``df'', ``s'', ``d'' and ``lr'' indicate using dataflow-guided retrieval, using sparse retrieval, using dense retrieval and using LLM reranker, respectively.
In other words, $\ours_{\text{df+s+d+lr}}$ is identical to $\ours_{\text{llmr}}$ shown in Tab.~\ref{tab:overall}.

From the results, we can observe that incorporating more retrieval paths brings improvements to code completion in most cases. 
Besides, using reranking module can significantly boost performance. 
These findings highlight the effectiveness of multi-path code retrieval and applying reranking in improving repository-level code completion.

\subsubsection{Comparisons between LLM Reranker and Distilled Reranker}
\label{sec:comp_reranker}

\begin{table*}[t]
  \centering
  \caption{Performance on ReccEval (Use 30\% data for evaluation). Bold and underlined values indicate the best and the second-best results, respectively.}
  \scalebox{0.63}{
    \begin{tabular}{ccccccccccccccccc}
    \toprule
    \multirow{3}[2]{*}{Methods} & \multicolumn{4}{c}{CodeGen-350M} & \multicolumn{4}{c}{SantaCoder-1.1B} & \multicolumn{4}{c}{StarCoder2-3B} & \multicolumn{4}{c}{Qwen2.5-Coder-7B} \\
          & \multicolumn{2}{c}{Code Match } & \multicolumn{2}{c}{Identifier Match} & \multicolumn{2}{c}{Code Match } & \multicolumn{2}{c}{Identifier Match} & \multicolumn{2}{c}{Code Match } & \multicolumn{2}{c}{Identifier Match} & \multicolumn{2}{c}{Code Match } & \multicolumn{2}{c}{Identifier Match} \\
          & EM    & ES    & EM    & F1    & EM    & ES    & EM    & F1    & EM    & ES    & EM    & F1    & EM    & ES    & EM    & F1 \\
    \midrule
    Zero-Shot & 4.46  & 38.09  & 9.33  & 26.24  & 6.87  & 42.52  & 13.22  & 30.28  & 8.30  & 44.81  & 13.99  & 33.17  & 12.20  & 47.55  & 18.20  & 36.36  \\
    CCFinder & 17.68  & 48.31  & 23.73  & 40.82  & 19.58  & 50.93  & 26.76 & 43.24  & 28.24  & 59.22  & 36.13  & 53.50  & 29.52  & 59.04  & 37.11  & 53.36  \\
    RG-1  & 17.63  & 48.94  & 24.07  & 39.54  & 24.04  & 54.80  & 31.06  & 46.44  & 29.63  & 58.96  & 35.73  & 51.81  & 33.52  & 61.59  & 39.67  & 54.93  \\
    RepoCoder & 20.35  & 50.84  & 27.21 & 42.78  & 26.86  & 56.79  & 34.24  & 49.22  & 35.21  & 63.14  & 41.82  & 57.26  & 34.65  & 62.19  & 41.16  & 55.52  \\ 
    DraCo & 22.71  & 51.83  & 29.98  & 46.40  & 30.19  & 59.70  & 39.21  & 55.66  & 36.60  & 64.78  & 45.41  & 61.76  & 39.98  & 66.23  & 48.18  & 62.82  \\
    RepoFuse & 21.58 & 51.54 & 28.81  & 44.70 & 29.98 & 58.60 & 37.62  & 52.71 & 34.03 & 62.24 & 41.62  & 56.94  & 39.11 & 65.54 & 46.03  & 60.13 \\
    \midrule 
    $\ours_{\text{llmr}}$ & \textbf{27.73} & \textbf{55.73} & \textbf{34.75} & \textbf{49.97} & \textbf{35.83} & \textbf{63.32} & \textbf{44.08} & \textbf{59.79} & \textbf{43.31} & \textbf{68.54} & \textbf{51.10} & \textbf{65.78}  & \textbf{47.57} & \textbf{70.82} & \textbf{54.84} & \textbf{68.38} \\
    $\ours_{\text{disr}}$ & \underline{23.58} & \underline{53.31} & \underline{30.55} & \underline{47.20} & \underline{32.65} & \underline{61.37} & \underline{41.16} & \underline{57.06} & \underline{39.88} & \underline{66.21} & \underline{47.62} & \underline{62.89}  & \underline{44.34} & \underline{68.42} & \underline{52.18} & \underline{64.47} \\
    \bottomrule
    \end{tabular}%
    }
  \label{tab:distilled_reranker}%
\end{table*}%

We further compare the results of using LLM reranker ($\ours_{\text{llmr}}$) and using distilled reranker ($\ours_{\text{disr}}$) on ReccEval in Tab.~\ref{tab:distilled_reranker}.
As $\ours_{\text{disr}}$ requires training, we randomly sample 70\% data for training and the remaining 30\% data is used for testing.
The results in Tab.~\ref{tab:distilled_reranker} are reported over the test data. 
From the results, we can observe that using the distilled reranker affects the performance compared to using LLM reranker, but the performance is still competitive.

\subsubsection{Comparisons among Different Retrieval Query Construction Methods.}

\begin{table}[t]
  \centering
  \caption{Comparisons among different query construction methods using sparse retrieval on ReccEval (Use 100\% data for evaluation).}
   \scalebox{0.75}{
    \begin{tabular}{ccccc}
    \toprule
    \multirow{3}[2]{*}{Methods} &  \multicolumn{4}{c}{Qwen2.5-Coder-7B} \\
          & \multicolumn{2}{c}{Code Match } & \multicolumn{2}{c}{Identifier Match} \\
          & EM    & ES  & EM    & F1 \\
    \midrule
    $\ours_{\text{jacc}}$ & 33.54  & 61.89  & 40.99  & 55.33 \\
    $\ours_{\text{last\_1}}$ & 32.08  & 61.04  & 39.58  & 54.91  \\
    $\ours_{\text{last\_3}}$ & 34.84  & 62.85  & 42.39  & 56.97  \\
    $\ours_{\text{last\_5}}$ & 34.21  & 62.26  & 41.50  & 56.04  \\
    $\ours_{\text{last\_10}}$ & 32.95  & 61.48  & 40.20  & 54.93  \\
    $\ours_{\text{s}}$ & 35.01  & 63.11  & 42.31  & 57.24  \\
    \bottomrule
    \end{tabular}%
    }
  \label{tab:query_compare}%
\end{table}%

In Tab.~\ref{tab:query_compare}, we compare the retrieval query construction method used in \ours and other retrieval query construction methods. 
$\ours_{\text{jacc}}$ represents constructing the retrieval query by selecting the chunk $c_i$ that is most similar to the target block (Alg.~\ref{alg:query_by_logit}) w.r.t. Jaccard similarity when using sparse retrieval. 
$\ours_{\text{last\_k}}$ means constructing the retrieval query with the last $k$ lines when using sparse retrieval. 
$\ours_{\text{s}}$ indicates using our retrieval query construction method in sparse retrieval.

From the results in Tab.~\ref{tab:query_compare}, we can observe that our retrieval query construction method significantly outperforms other simple methods.

\subsubsection{Computational Cost}
\label{sec:cost}

\begin{table}[t]
\centering
\caption{Average cost (second) for a query on ReccEval, excluding the cost for code generation of code LLM.}
\scalebox{0.85}{
\begin{tabular}{@{}cccc@{}}
\toprule
RepoCoder & DraCo & RepoFuse & \ours \\ \midrule
0.21     & 0.04 & 0.15  & 0.23                  \\ \bottomrule
\end{tabular}
}
\label{tab:cost}%
\end{table}

\begin{table}[t]
\centering
\caption{Average cost (second) for each step of \ours on ReccEval.}
\scalebox{0.85}{
\begin{tabular}{@{}ccccc@{}}
\toprule
\multirow{2}{*}{\begin{tabular}[c]{@{}c@{}}QueryQuery\\ Construction\end{tabular}} & \multicolumn{3}{c}{Multi-path Retrieval} & \multirow{2}{*}{\begin{tabular}[c]{@{}c@{}}Distilled\\ Reranker\end{tabular}} \\
                                                                                   & Sparse      & Dense      & Dataflow      &                                                                               \\ \midrule
\multicolumn{1}{c}{0.14}                                                           & 0.002       & 0.015      & 0.03          & 0.06                                                                          \\ \bottomrule
\end{tabular}
}
\label{tab:cost2}%
\end{table}

Tab.~\ref{tab:cost} provides the average cost for a query on ReccEval, excluding the time for code generation of code LLM. 
We can see that DraCo is the fastest since it only uses a single-way, dataflow-guided retrieval, and no similarity search is needed. 
Other methods are slower than DraCo as they all require similarity search. 
\ours incurs higher cost than baselines since it uses more ways of retrieval, applies log probability guided probing and a distilled reranker from \textsc{BestFit} code reranking.
It indeed involves more steps than baselines. 
Nevertheless, the cost of \ours is close to RepoCoder and RepoFuse even though we do not implement \ours using advanced acceleration methods.
Considering the significant improvements of \ours over baselines, we believe the subtle increase of overhead is acceptable.

Tab.~\ref{tab:cost2} reports the average cost for each step of \ours on ReccEval. 
We can see that the largest cost comes from query construction. 
Note that, for multi-path code retrieval, the cost is decided by the slowest path (i.e., dataflow-guided retrieval).

Since the main goal of this work is not accelerating code completion and we do not apply advanced acceleration techniques, we believe the cost of \ours can be further reduced and it will not hinder the practical use of \ours. 
We suggest some possible directions, including constructing retrieval query in batches and parallelization, accelerating log probability guided probing through applying an inference speedup framework (e.g., vLLM\footnote{\url{https://github.com/vllm-project/vllm}}) on the prober LLM, using fast sparse and dense retrieval libraries (e.g., Faiss\footnote{\url{https://github.com/facebookresearch/faiss}}) and distilling LLM reranker into a much smaller distilled reranker.

\subsubsection{Impacts of Numbers of Retained Reranked Code Knowledge Pieces}
\label{sec:impact_retained_number}

\begin{figure}[t]
    \centering
    \includegraphics[width=1\linewidth]{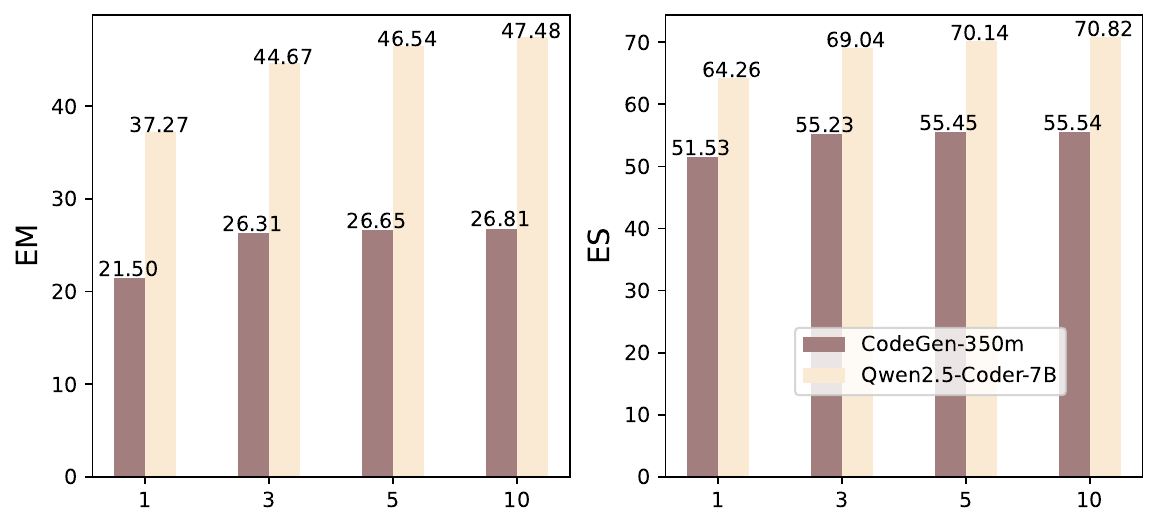}
    \caption{Impacts of numbers of retained reranked code knowledge on ReccEval (Use 100\% data for evaluation).}
    \label{fig:rerank_topK}
\end{figure}

The reranking step in \ours only retains top-$u$ pieces of relevant code knowledge. 
In Fig.~\ref{fig:rerank_topK}, we report the different results w.r.t. EM and ES (code match) when setting $u$ to 1, 3, 5 and 10 on ReccEval (Use 100\% data for evaluation). 
From the results, we can observe that: 
\begin{itemize}[leftmargin=10pt,topsep=1pt,itemsep=0.2pt] 
    \item Regardless of the used code LLM, using a larger $u$ can enhance the quality of code completion. 

    \item The improvements are noticeable when $u$ is increased from a small value (e.g., increase from 1 to 3), but the enhancement becomes marginal when $u$ further grows (e.g., increase from 5 to 10). The diminishing improvement suggests that larger $u$ may potentially introduce less relevant and necessary code knowledge that does not significantly contribute to code completion.
\end{itemize}

\section{Human Evaluation on Ranking Quality}
\label{app:human}

We invite three students in Computer Science major to manually verify the results of LLM reranking.

On ReccEval, we randomly sample 100 code completion cases and deploy our LLM reranking strategy (Sec.~\ref{sec:align}) to rerank retrieved code knowledge.
We ask three student volunteers who are master students in computer science to independently rate the 100 reranking lists using a score between 1 and 5 with 5 indicating the best quality.
They are informed that the results will be used in a research paper.
Finally, the average scores given by the three students are 3.69, 3.76 and 4.03, respectively.
From the results of human evaluation, we can conclude that the LLM reranker indeed has a positive effect and sorts retrieved code knowledge towards the right order.

\section{Related Work}

Recently, there is a surge of works on repository-level code completion. 
RepoFusion~\cite{abs-2306-10998} models the entire repository structure for context-aware completion. RepoCoder~\cite{RepoCoder} employs an iterative code retrieval and generation mechanism.
CoCoMIC~\cite{DingWARNBRX24} enhances accuracy by combining cross-file and intra-file contexts.
RepoFormer~\cite{0054AZR024} fine-tunes models to dynamically decide context retrieval needs. 
GraphCoder~\cite{GraphCoder} models control-flow dependency, data dependency, and control dependency to construct code knowledge graphs for code completion. 
ProCC~\cite{abs-2405-07530} integrates prompt engineering with contextual bandits for multi-perspective code completion.

DraCo~\cite{Dataflow} and RepoFuse~\cite{abs-2402-14323} are closely related to \ours.
DraCo retrieves dataflow-guided information to augment the code completion prompt.
RepoFuse~\cite{abs-2402-14323} fuses analogy context and rationale context, and uses a code LM like UniXcoder~\cite{GuoLDW0022} to choose the most similar chunks to the target chunk to construct code completion prompt. 
Despite their accuracy, they do not fully address problems discussed in Sec.~\ref{sec:intro}.

\section{Conclusion}

This paper introduces \ours, a novel framework for repository-level code completion. 
Its core parts include the log probability guided query construction, a multi-path code retrieval mechanism, and finding necessary code knowledge through preference-aligned reranking. 
Experimental results demonstrate that \ours significantly and consistently outperforms state-of-the-art methods on benchmark data.

In the future, we plan to design new strategies such as joint training of code retriever and code LLM to further alleviate the misalignment between retrieval knowledge and necessary knowledge for better repository-level code completion.
We also plan to construct a new benchmark to evaluate the generalization of \ours.

\section*{Limitations}

We address the misalignment between code retriever and code LLM from the perspective of designing rerankers and modifying retrieval results while code LLM is not updated accordingly. This idea may not fully alleviate the misalignment, and future work may involve designing new strategies, such as joint training or more efficient interfacing mechanisms for both code retrieval and code LLM.

\section*{Acknowledgments}
This work was supported by National Natural Science Foundation of China (No. 62572410, 42171456), Natural Science Foundation of Xiamen, China (No. 3502Z202471028) and Ant Group through CCF-Ant Research Fund.

\bibliography{main}

\appendix

\section{Results on CCEval}
\label{app:crosscodeeval}

In addition to ReccEval, we also conduct experiments on the CCEval benchmark.
Tab.~\ref{tab:crosscodeeval_100} reports the results on CCEval using LLM reranker in \ours and 100\% data for evaluation.
Tab.~\ref{tab:crosscodeeval_30} provides the results on CCEval using distilled reranker in \ours and 30\% data for evaluation.  
Tab.~\ref{tab:crosscodeeval_repoformer} shows the results using Repoformer-3B with RG-1 as the retriever on CCEval.

From Tab.~\ref{tab:crosscodeeval_100} and Tab.~\ref{tab:crosscodeeval_30}, we can see similar trends as observed on ReccEval. 
From Tab.~\ref{tab:crosscodeeval_repoformer}, we can observe that the performance of Repoformer-3B using both left and right contexts gets significantly improved compared to Repoformer-3B using only left context since it is optimized to use left and right parts of the cursor position.
By comparing $\ours_{\text{llmr}}$ with StarCoder2-3B in Tab.~\ref{tab:crosscodeeval_100} and Repoformer-3B in Tab.~\ref{tab:crosscodeeval_repoformer}, we can observe that $\ours_{\text{llmr}}$ which only considers left context outperforms Repoformer-3B that uses both left and right contexts, when 3B code LLM is used as code generator, showing the effectiveness of \ours.

\begin{table*}[!t]
  \centering 
  \caption{Performance on CCEval (Use 100\% data for evaluation). Bold and underlined values indicate the best and the second-best results, respectively.} 
  \scalebox{0.65}{
    \begin{tabular}{ccccccccccccccccc}
    \toprule
    \multirow{3}[2]{*}{Methods} & \multicolumn{4}{c}{CodeGen-350M} & \multicolumn{4}{c}{SantaCoder-1.1B} & \multicolumn{4}{c}{StarCoder2-3B} & \multicolumn{4}{c}{Qwen2.5-Coder-7B} \\
          & \multicolumn{2}{c}{Code Match } & \multicolumn{2}{c}{Identifier Match} & \multicolumn{2}{c}{Code Match } & \multicolumn{2}{c}{Identifier Match} & \multicolumn{2}{c}{Code Match } & \multicolumn{2}{c}{Identifier Match} & \multicolumn{2}{c}{Code Match } & \multicolumn{2}{c}{Identifier Match} \\
          & EM    & ES    & EM    & F1    & EM    & ES    & EM    & F1    & EM    & ES    & EM    & F1    & EM    & ES    & EM    & F1 \\
    \midrule
    Zero-Shot & 2.70  & 43.02  & 8.26  & 37.85  & 4.35  & 46.52  & 10.58  & 41.88  & 6.53  & 48.56  & 12.91  & 44.52  & 11.11  & 52.19  & 18.09  & 48.35  \\
    CCFinder & 10.58  & 48.65  & 17.19  & 45.96  & 14.63  & 53.44  & 23.08  & 51.90  & 21.08  & 58.11  & 29.42  & 57.46  & 24.80  & 59.52  & 33.47  & 62.61  \\
    RG-1  & 8.78  & 49.54  & 16.47  & 46.33  & 12.83  & 53.78  & 21.99  & 51.76  & 17.78  & 57.74  & 27.35  & 56.55  & 22.51  & 61.84  & 32.91  & 60.68  \\
    RepoCoder & 10.58  & \underline{51.07}  & 19.06  & 48.93  & 15.12  & 55.66  & 24.62  & 53.78  & 21.24  & 60.90  & 31.56  & 60.00  & 25.89  & 63.65  & 36.21  & 63.09  \\ 
    DraCo & \underline{12.83}  & 50.71  & \underline{20.33}  & \underline{48.91}  & \underline{19.70}  & \underline{57.17}  & \underline{29.04}  & \underline{56.82}  & \underline{26.68}  & \underline{62.11}  & \underline{36.29}  & \underline{62.75}  & \underline{30.69}  & \underline{65.46}  & \underline{40.64}  & \underline{66.26}  \\
    RepoFuse & 11.22 & 50.78 & 19.29  & 48.77 & 17.64 & 56.52 & 27.02  & 55.65 & 23.34 & 61.28 & 33.43  & 61.16  & 27.73 & 64.76 & 38.39  & 64.82 \\
    \midrule
    $\ours_{\text{llmr}}$ & \textbf{14.11}  & \textbf{52.44}  & \textbf{22.28}  & \textbf{51.56}  & \textbf{22.89} & \textbf{59.92}  & \textbf{32.42} & \textbf{60.24}  & \textbf{30.66}  & \textbf{65.46}  & \textbf{41.13}  & \textbf{66.62}  & \textbf{35.20}  & \textbf{68.93}  & \textbf{45.97}  & \textbf{70.47}  \\
    \bottomrule
    \end{tabular}%
    }
  \label{tab:crosscodeeval_100}%
\end{table*}%

\begin{table*}[!t]
  \centering 
  \caption{Performance on CCEval (Use 30\% data for evaluation). Bold and underlined values indicate the best and the second-best results, respectively.}
  \scalebox{0.65}{
    \begin{tabular}{ccccccccccccccccc}
    \toprule
    \multirow{3}[2]{*}{} & \multicolumn{4}{c}{CodeGen-350M} & \multicolumn{4}{c}{SantaCoder-1.1B} & \multicolumn{4}{c}{StarCoder2-3B} & \multicolumn{4}{c}{Qwen2.5-Coder-7B} \\
          & \multicolumn{2}{c}{Code Match } & \multicolumn{2}{c}{Identifier Match} & \multicolumn{2}{c}{Code Match } & \multicolumn{2}{c}{Identifier Match} & \multicolumn{2}{c}{Code Match } & \multicolumn{2}{c}{Identifier Match} & \multicolumn{2}{c}{Code Match } & \multicolumn{2}{c}{Identifier Match} \\
          & EM    & ES    & EM    & F1    & EM    & ES    & EM    & F1    & EM    & ES    & EM    & F1    & EM    & ES    & EM    & F1 \\
    \midrule
    Zero-Shot & 2.57  & 41.34  & 6.77  & 35.29  & 3.88  & 45.72  & 9.41  & 40.11  & 6.88  & 47.45  & 11.79  & 42.26  & 9.65  & 50.89  & 15.29  & 45.10  \\
    CCFinder & 9.33  & 46.20  & 14.24  & 42.69  & 13.41  & 52.06  & 21.18  & 50.08  & 20.30  & 56.07  & 25.90  & 53.32  & 23.53  & 58.82  & 30.00  & 59.85  \\
    RG-1  & 8.52  & 48.11  & 15.52  & 43.90  & 11.76  & 53.03  & 20.24  & 49.37  & 16.57  & 54.80  & 23.80  & 51.89  & 21.53  & 60.49  & 31.06  & 57.98  \\
    RepoCoder & 10.62  & \underline{49.49}  & 17.39  & 46.22  & 14.59  & 54.87  & 23.06  & 51.74  & 21.24  & 58.46  & 28.47  & 56.28  & 24.71  & 62.35  & 33.88  & 60.57  \\ 
    DraCo & \underline{11.55}  & 47.71  & 17.04  & 44.95  & \underline{19.60}  & 55.62  & \underline{26.84}  & 53.52  & \underline{25.67}  & 59.59  & 32.32  & 58.68  & \textbf{29.99}  & \underline{63.88}  & \underline{36.17}  & \textbf{63.86}  \\
    RepoFuse & 11.41 & \textbf{49.87} & \textbf{18.82}  & \textbf{46.40} & 18.35 & \underline{56.00} & 26.59  & \underline{54.00} & 23.06 & \textbf{61.27} & \underline{32.59}  & \underline{59.56}  & 26.71 & 63.36 & 35.88  & 62.08 \\
    \midrule
    $\ours_{\text{disr}}$ & \textbf{11.78}  & 48.39  & \underline{17.74}  & \underline{45.04}  & \textbf{20.07} & \textbf{56.30} & \textbf{27.77} & \textbf{54.75}  & \textbf{26.25} & \underline{60.53}  & \textbf{33.49} & \textbf{60.21} & \underline{28.70} & \textbf{64.56}  & \textbf{36.52}  & \underline{63.76}  \\
    \bottomrule
    \end{tabular}%
    }
  \label{tab:crosscodeeval_30}%
\end{table*}%

\begin{table}[H]
  \centering
  \caption{Results of RepoFormer-3B on CCEval (Use 100\% Data for evaluation). $l$: only use left context. $lr$: use left and right contexts.}
  \scalebox{0.8}{
    \begin{tabular}{ccccc}
    \toprule
    \multirow{3}[1]{*}{Dataset} & \multicolumn{4}{c}{RepoFormer-3B} \\
          & \multicolumn{2}{c}{Code Match } & \multicolumn{2}{c}{Identifier Match} \\
          & EM    & ES    & EM    & F1 \\
    \midrule 
    $\text{CCEval}_{l}$ & 8.18  & 50.19  & 15.53  & 46.25  \\
    $\text{CCEval}_{lr}$ & 25.29  & 63.45  & 33.77  & 61.48  \\
    \bottomrule
    \end{tabular}%
    }
  \label{tab:crosscodeeval_repoformer}%
\end{table}%

\end{document}